\pdfoutput=1
\documentclass[11pt]{article}

\usepackage{ACL2023}
\usepackage{times}
\usepackage{latexsym}

\usepackage[T1]{fontenc}
\usepackage[utf8]{inputenc}

\usepackage{microtype}

\usepackage{inconsolata}

\usepackage{soul}
\usepackage{booktabs}
\usepackage{enumitem}
\usepackage{graphicx}
\usepackage{multirow}
\usepackage{booktabs}
\usepackage{amssymb}
\usepackage{amsmath}

\newcommand{\nop}[1]{}

\newcommand{\temp}[1]{\textcolor{teal}{#1}}
\title{Learning to Simulate Natural Language Feedback for\\Interactive Semantic Parsing}

\author{Hao Yan\textsuperscript{1}, Saurabh Srivastava\textsuperscript{1}, Yintao Tai\textsuperscript{2\thanks{\hspace{.07in}Yintao Tai was a remote intern at GMU during this project.}} , Sida I. Wang\textsuperscript{3}, Wen-tau Yih\textsuperscript{3}, Ziyu Yao\textsuperscript{1}\\
        \textsuperscript{1}George Mason University, \textsuperscript{2}The University of Edinburgh, \textsuperscript{3}Meta AI\\
        \textsuperscript{1}{\tt \{hyan5, ssrivas6, ziyuyao\}@gmu.edu}, \textsuperscript{2}{\tt y.tai-6@sms.ed.ac.uk}\\
        \textsuperscript{3}{\tt \{sida, scottyih\}@meta.com}\\
}
\begin{document}
\maketitle

\begin{abstract}
\emph{Interactive semantic parsing based on natural language (NL) feedback}, where users provide feedback to correct the parser mistakes, has emerged as a more practical scenario than the traditional one-shot semantic parsing. However, prior work has heavily relied on human-annotated feedback data to train the interactive semantic parser, which is prohibitively expensive and not scalable.
In this work, we propose a new task of \emph{simulating NL feedback for interactive semantic parsing}. 
{We accompany the task with a novel feedback evaluator. The evaluator is specifically designed to assess the quality of the simulated feedback, based on which we decide the best feedback simulator from our proposed variants.}
On a text-to-SQL dataset, we show that our feedback simulator can generate high-quality NL feedback to boost the error correction ability of a specific parser. In low-data settings, our feedback simulator can help achieve comparable error correction performance as trained using the costly, full set of human annotations.\footnote{Our code is publicly available at \url{https://github.com/hyan5/Learning_to_Simulate_NL_Feedback}.}

\end{abstract}

\section{Introduction}\label{sec:intro}
The state of NLP research has long been dominated by training and evaluating \emph{single-turn} models, which, given a task input, produce the output and terminate the task immediately. However, in the more practical scenario of NLP applications (e.g., smart-home virtual assistance), users often anticipate \emph{multi-turn} interactions, such as being able to provide \emph{feedback} to the model output \cite{de2020towards}. In doing this, not only can the model obtain more information and guidance to improve its task performance, but it also provides human users a mechanism to intervene in the model decision-making for safety purposes. However, training a neural model to understand human feedback requires a large number of human annotations, which has hindered the advancement of this line of research.

\begin{figure}
    \centering
    \includegraphics[width=\linewidth]{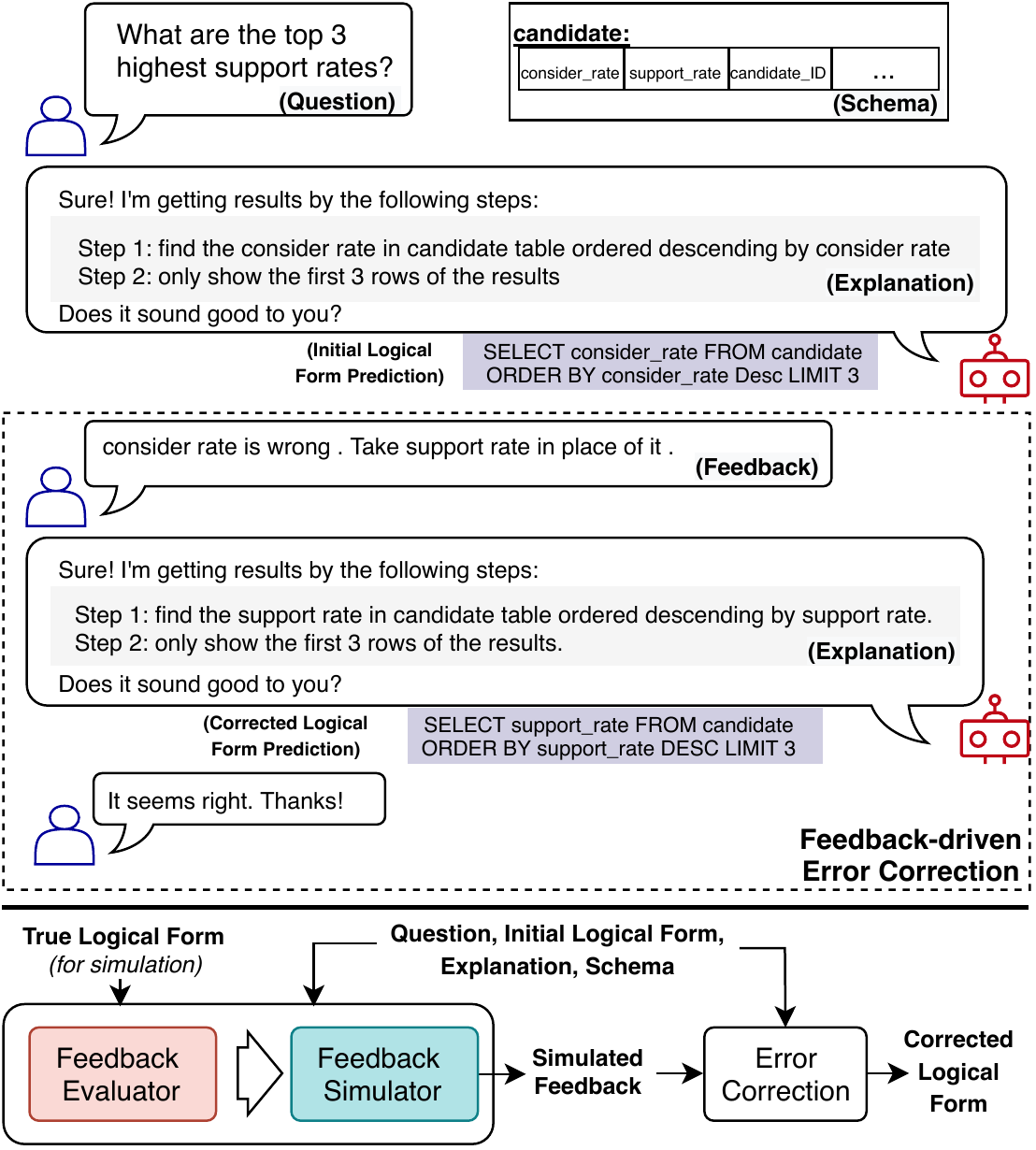}
    \caption{Illustration of interactive semantic parsing, where the parser solicits NL feedback for error correction (example based on text-to-SQL). In this work, we aim to simulate such NL feedback at scale to facilitate the error correction model training. To this end, we proposed a feedback evaluator for promoting this task, and experiment with different feedback simulator variants.}
    \label{fig:overview}
\end{figure}

In this paper, we investigate this problem under semantic parsing. Semantic parsing is the task of translating NL sentences into their formal meaning representations (i.e., logical forms), which has been adopted for applications such as question answering \cite{reddy-etal-2014-large, dong-lapata-2016-language, yu-etal-2018-spider, gu2021beyond} and dialogue systems \cite{gupta-etal-2018-semantic-parsing, andreas-etal-2020-task, cheng-etal-2020-conversational}.
The pressing need for further improving its application performance has motivated the research of interactive semantic parsing, where a semantic parser presents its parsing results to the user and requests user feedback for error correction \cite{gur-etal-2018-dialsql, yao-etal-2019-model, li-etal-2020-mean, elgohary-etal-2020-speak}. In this work, we follow \citet{labutov-etal-2018-learning, elgohary-etal-2020-speak} to consider \emph{NL feedback}, i.e., a sentence describing which parts of the generated logical form contain errors and how to correct them. We illustrate this paradigm in Figure~\ref{fig:overview}.

Despite its promise, prior work has heavily relied on human-annotated feedback data to train the error correction model. For example, \citet{elgohary-etal-2020-speak} deployed the Seq2Struct parser \cite{shinSeq2Struct} and recruited 10 crowd workers to provide feedback annotations, which has been shown to be both costly and time-consuming (6 minutes per annotation as reported). Moreover, since this feedback collection procedure is bound to a specific parser, the collected feedback may not generalize well to resolving errors made by different parsers.

Motivated by the above observations, in this paper, we propose the task of \emph{simulating NL feedback for interactive semantic parsing}. Specifically, given the initial user command, a model-generated incorrect logical form, the ground-truth logical form for the simulation purpose, as well as other contextual information, the goal is to generate an NL feedback sentence encoding the error correction information in a way that is close to the real-user feedback. We assume a small set of human-annotated feedback to bootstrap this task, but aim for an effective feedback simulator that can further simulate feedback for different semantic parsers at scale. While prior work has attempted a similar task \cite{yao2019interactive,elgohary-etal-2021-nl,mo-etal-2022-towards}, none of them carefully defined the task (e.g., how to evaluate simulated feedback) and investigated advanced simulation methods.

To facilitate this research, we first propose a feedback evaluator that can be used to assess different simulators. In particular, our feedback evaluator is designed to evaluate whether the simulated feedback is \emph{logically consistent} with the user error correction intent, a critical attribute that cannot be achieved by existing text evaluation metrics \cite{papineni-etal-2002-bleu, zhang2019bertscore}. Instead of comparing the simulated feedback with the human-annotated one, we propose to compare it with the \emph{template feedback}, which is not only logic-wisely less noisy but also scalable to cases when human annotations are not available. Human evaluation shows that our feedback evaluator can more precisely assess the simulated feedback. We also propose a set of feedback simulators based on the pre-trained T5 model \cite{raffel2020exploring}, and decide the best using our evaluator.

To demonstrate the advantages of our feedback simulator, we conduct experiments on SPLASH \cite{elgohary-etal-2020-speak}, a dataset containing human-annotated feedback to mistakes of the Seq2Struct parser \cite{shinSeq2Struct} in text-to-SQL semantic parsing \cite{yu-etal-2018-spider}. 
We first show that our feedback simulator trained on SPLASH can be used to simulate NL feedback for a different parser, using EditSQL \cite{zhang-etal-2019-editing} as an example. The resulting simulated feedback, when being used to augment the SPLASH training set, leads to improved error correction performance for both Seq2Struct and particularly EditSQL. We further demonstrate that even in the low-data setting (i.e., using a small portion of SPLASH), our feedback simulator can still produce high-quality NL feedback, based on which we can train the error correction model to a comparable performance level as its counterpart trained using the full SPLASH. This implies that our feedback simulator can be very helpful when there are limited annotation budgets.
\section{Simulating Natural Language Feedback for Interactive Semantic Parsing}\label{sec:method}

\subsection{Overview}
We illustrate the scenario of interactive semantic parsing in Figure~\ref{fig:overview}. Given an initial user question $Q$, as well as other contextual information (e.g., database schema in text-to-SQL semantic parsing, denoted as $S$), the semantic parser will first produce an initial logical form $Y_{init}$. The semantic parser will then present a logical form explanation $E$ to the user.\footnote{We assume that the user is not professional in understanding and writing the logical form (otherwise they would not need to use the parser). Therefore, each logical form is presented to the user via an explanation. In practice, we implement the explanation via NL templates following \citet{elgohary-etal-2020-speak}, whereas leaving the exploration of more advanced explanation methods to the future.}
After receiving the explanation, the user is prompted to give an NL feedback sentence $F$, describing which parts of the logical form $Y_{init}$ contain errors and how to correct them. This information is perceived by the error correction model of the interactive semantic parser to refresh its logical form prediction, hoping that the new prediction $Y_{fix}$ can be the same as the ground truth $Y^*$.

Training the interactive semantic parser (or more precisely, its error correction model) to understand NL feedback requires abundant human-annotated feedback data. In this work, we propose a new task of \emph{simulating NL feedback for interactive semantic parsing}, aiming to reduce the reliance on human annotations. We assume a set of human-annotated feedback data $\mathcal{D}_{train}$, consisting of tuples of $(Q, S, Y_{init}, E, F, Y^*)$, to bootstrap such a feedback simulator, but aim for an effective simulator that can generate high-quality NL feedback at scale. The simulated feedback can then be used to assist the error correction model training.

To facilitate this task, we first introduce a feedback evaluator in Section~\ref{subsec:feedback_eval}, and then present a set of feedback simulators in Section~\ref{subsec:feedback_gen}.

\subsection{Feedback Evaluation} \label{subsec:feedback_eval}

It is critical that the simulated feedback is both \emph{fluent} (i.e., as how real users speak) and \emph{logically consistent} with the user error correction intent (i.e., precisely articulating which parts of the predicted logical form are wrong and how to correct them). While the prevalent use of pre-trained language models has been able to improve generation fluency dramatically \cite{radford2019language, lewis-etal-2020-bart, raffel2020exploring}, ensuring that the simulated feedback has a consistent logic with the simulation intent is still a challenging problem. 
This motivates us to accompany the feedback simulation task with an evaluator that can be reused by future researchers to assess the quality of the simulated feedback from a logical front. To this end, we design a feedback evaluator as elaborated below. The evaluator will be trained using the available feedback annotations $\mathcal{D}_{train}$.

\subsubsection{Task Formulation \& Architecture}
\begin{figure}[t!]
    \centering
    \includegraphics[width=\linewidth]{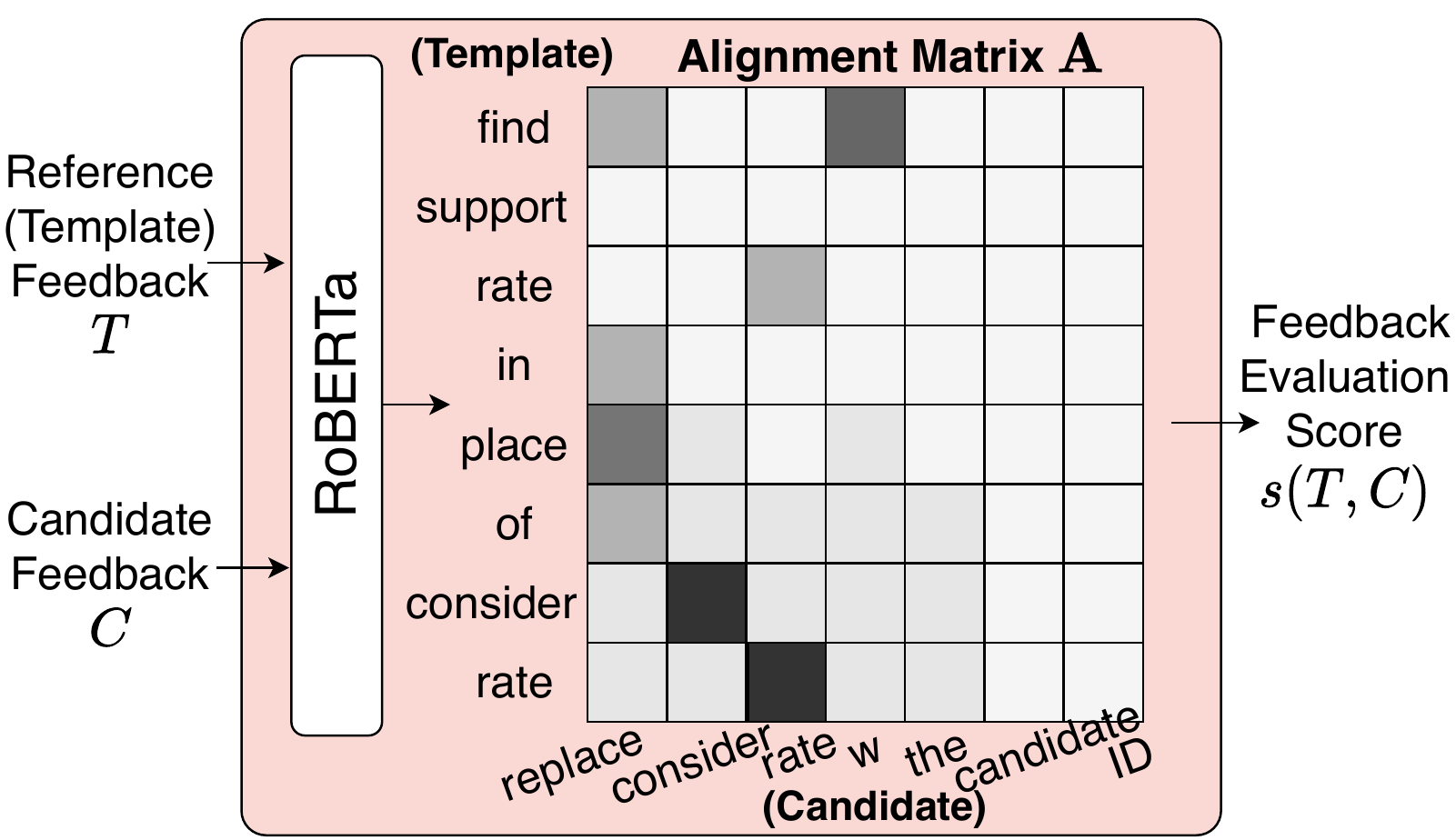}
    \caption{Our feedback evaluator assesses the logical quality of the simulated feedback by leveraging the template feedback as reference.}
    \vspace{-5mm}
    \label{fig:feedback_eval}
\end{figure}

Without the loss of generality, given a reference feedback sentence $T=(t_1, t_2, ..., t_N)$ and a candidate feedback sentence $C=(c_1, c_2, ..., c_M)$, the goal of a feedback evaluator is to produce a score $s(T, C)$, such that when the candidate $C$ is logically consistent with the error correction intent (as reflected in the reference $T$), the evaluator predicts a high score $s$, and vice versa. In our task, the candidate $C$ is the simulated NL feedback. As for the reference $T$, instead of using the human-annotated feedback, we use a \emph{template feedback} derived from the same context. A simplified example is shown in Figure~\ref{fig:feedback_eval}, which describes the column replacement in text-to-SQL parsing using a template ``\texttt{find [Col$_{correct}$] in place of [Col$_{wrong}$]}'', where ``\texttt{[Col$_{correct}$]}'' and ``\texttt{[Col$_{wrong}$]}'' are placeholders for correct and incorrect columns, respectively. We include more details of our templates in Appendix~\ref{app:subsec:template}. Using template feedback as reference offers two advantages. First, it provides a cleaner standard than the human-annotated one, which we empirically found to contain inaccurate or incomplete error descriptions. Second, since template feedback can be generated automatically, it can easily scale to cases when human annotations are not available.

In order to capture the feedback semantics at the logical level, we adopt a model architecture similar to that of \citet{zhang2019bertscore}, which first computes the token-level similarity between the candidate and the reference, and then aggregates the information toward scoring their similarity at the sentence level (Figure~\ref{fig:feedback_eval}). Specifically, the model takes the candidate $C$ and the reference $T$ as input and first obtains their token-level contextual representations via RoBERTa \cite{RoBERTa}, obtaining $\mathbf{h_n^T}, \mathbf{h_m^C} \in \mathbb{R}^d$, where $d$ is the embedding size, for token $t_n$ ($n$=1 ,..., $N$) and $c_m$ ($m$=1 ,..., $M$), respectively. We then obtain a token-level similarity matrix $\mathrm{\mathbf{A}}\in\mathbb{R}^{N\times M}$ by calculating the cosine similarity between every pair of tokens in the reference and the candidate, i.e., $\mathrm{\mathbf{A}}_{nm}=\frac{\mathbf{h_n^T}^\top\cdot \mathbf{h_m^C}}{||\mathbf{h_n^T}||\cdot ||\mathbf{h_m^C}||}$ .

The sentence-level similarity between the reference and the candidate can then be derived from their token-level similarities. We notice that not only should the candidate align with the reference (precision) but the alignment should also hold in the opposite direction (recall). Therefore, our sentence-level similarity first calculates the precision and the recall between the two sentences, i.e., $s_{prec}(T,C) = \frac{1}{M}\sum_{m=1}^{M}\max_n\mathrm{\mathbf{A}}_{nm}$, $s_{recall}(T,C) = \frac{1}{N}\sum_{n=1}^{N}\max_m\mathrm{\mathbf{A}}_{nm}$, and then calculates their average as the final score, i.e., $s(T,C) = \frac{1}{2}(s_{prec}+s_{recall})$.

We train the evaluator to contrast positive $C_{pos}$ and negative $C_{neg}$ candidates via a hinge loss:
\begin{align*}
    \mathcal{L}^{margin} &= \max(0, m - s(T,C_{pos}) + s(T,C_{neg}))\\
    & + \lambda (|\mathrm{\mathbf{A}}_{pos}|_1 + |\mathrm{\mathbf{A}}_{neg}|_1)
\end{align*}
where $m$ is the margin, $|\mathrm{\mathbf{A}}|_1$ denotes the L1 norm encouraging sparse alignments, and $\lambda$ is the weight factor.
In practice, we will use the human-annotated feedback $F$ as the positive candidate and the negative one will be introduced shortly.

\vspace{1mm}
\noindent \textbf{Supervision on Token-level Alignment.}
Inspired by \citet{yin-etal-2021-compositional}, we additionally introduce alignment supervision on tokens that can be derived from task-specific information. For example, in the task of text-to-SQL semantic parsing, it is easy to derive schema items appearing in the template feedback, and their correspondences in the human-annotated feedback can be extracted using fuzzy string matching \cite{lin-etal-2020-bridging}. This results in a prior alignment matrix, denoted as $\mathrm{\mathbf{A}}^{prior} \in \mathbb{R}^{N \times M}$ in our work. Specifically, every element in the matrix is set to 1 if the corresponding tokens in the reference and the candidate should be aligned, and 0 otherwise. The supervision is realized by the loss:
\[\mathcal{L}^{prior}=\sum_{n=1}^N\sum_{m=1}^M(\mathrm{\mathbf{A}}_{nm}-\mathrm{\mathbf{A}}_{nm}^{prior})^2\times \mathrm{\mathbf{A}}_{nm}^{mask},
\]
where $\mathrm{\mathbf{A}}^{mask} \in \mathbb{R}^{N \times M}$ is a mask matrix used to eliminate the impact of the supervision on tokens for which we cannot derive their correct alignments. Specifically, for tokens in the same row or column as those aligned tokens, we assign $\mathrm{\mathbf{A}}^{mask}_{nm}$ to 1 for them, and 0 otherwise.
% ; in this case, we set $\mathrm{\mathbf{A}}^{mask}_{nm}$ as 0.
The final loss function for training the evaluator is:
\[
    \mathcal{L}=\mathcal{L}^{margin}+\gamma \mathcal{L}^{prior},
\]
where $\gamma$ is the weight of the prior loss.

\vspace{1mm}
\noindent \textbf{Negative Candidate Feedback.}
Motivated by the observation that most feedback is about correcting certain values and schema items (e.g., table and column names in text-to-SQL parsing), we sample negative feedback from the human-annotated feedback by replacing their values and schema items with random ones. Taking text-to-SQL semantic parsing as an example, we replace the column name ``location description'' in the feedback ``use location name instead of \emph{location description}'' with a different column in the same database, such as ``document type description'', resulting in a negative feedback sentence ``use location name instead of \emph{document type description}''. In this way, our feedback evaluator will be trained to capture such subtle differences between good and bad feedback.

\vspace{1mm}
\noindent \textbf{Post-processing.} 
To further encourage one-to-one alignments between the reference and the candidate, we follow \citet{li-etal-2020-mean} to perform Bipartite Matching at inference time. Furthermore, we noticed that spans in the reference (i.e., template) feedback contribute differently to describing the error correction intent. For example, when a user would like to replace a certain schema item with an alternative one, they will indicate the correct alternative, but may or may not mention the incorrect one. Therefore, we additionally weigh different spans in the reference feedback while calculating the similarity score. More details are shown in Appendix~\ref{app:subsec:postprocessing}.
\subsection{Feedback Simulation}\label{subsec:feedback_gen}
Given the initial user question $Q$, the initial logical form prediction $Y_{init}$, the gold logical form $Y^*$ (for the simulation purpose), as well as other information such as the explanation $E$ and the context $S$, a feedback simulator aims to produce a feedback sentence $F$ that is similar to how humans give corrective instructions to the semantic parser. 

\begin{figure}[t!]
    \centering
    \includegraphics[width=\linewidth]{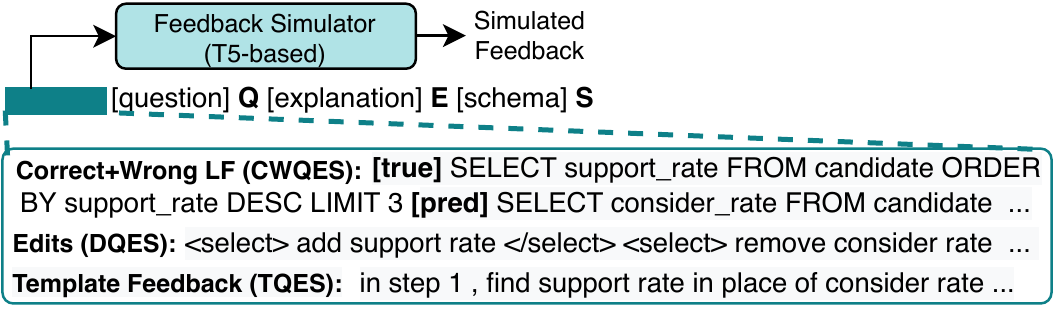}
    \caption{Our feedback simulator variants with different ways of error correction intent representations.}
    \vspace{-5mm}
    \label{fig:feedback_sim}
\end{figure}

In this section, we present three variants of feedback simulator, all based on fine-tuning the pre-trained T5 model \cite{raffel2020exploring}. The variants are only different in the way how they represent the error correction intent. Figure~\ref{fig:feedback_sim} gives an overview of them. \textbf{(1) \ul{CW}QES}: In this variant, we simply include the \ul{C}orrect and \ul{W}rong logical forms as input and train the model to simulate feedback.
\textbf{(2) \ul{D}QES}: Inspired by \citet{elgohary-etal-2021-nl}, we also explore feeding the e\ul{D}its of revising the incorrect logical form $Y_{init}$ into the gold one $Y^*$ as input. Compared with feeding the raw logical forms, this variant will make the simulation task easier, because, unlike the former, the simulator will have no need to understand the two logical forms and infer their differences. In practice, we follow \citet{elgohary-etal-2021-nl} and represent the edits in a linearized form.
\textbf{(3) \ul{T}QES}: Finally, we propose to represent the edits using their \ul{T}emplate description, which is the same as our template feedback introduced in Section~\ref{subsec:feedback_eval}. In this way, the task of feedback simulation can be viewed as paraphrasing the template feedback and making it more similar to how the real user speaks. The advantage of this variant lies in that it can better unlock the power of language models pre-trained on textual data (e.g., T5), when the program-liked edits are replaced by their textual descriptions. Same as the feedback evaluator, our feedback simulator will be trained on the available human annotations $\mathcal{D}_{train}$.
\section{Experiments}

\begin{table*}[ht!]
    \centering\small
    \resizebox{\linewidth}{!}{%
    \begin{tabular}{l | ccccc | ccccc}
    \toprule
    \multirow{3}{*}{\textbf{Model}} & \multicolumn{5}{c|}{\textbf{SPLASH-Test}} & \multicolumn{5}{c}{\textbf{EditSQL-Test}} \\
    \cmidrule{2-11}
    & \textbf{Corr Acc.} & \textbf{Progress}& \textbf{Edit-Dec} & \textbf{Edit-Inc} & \textbf{E2E} & \textbf{Corr Acc.} & \textbf{Progress}& \textbf{Edit-Dec} & \textbf{Edit-Inc} & \textbf{E2E}\\
    & ($\uparrow$) & ($\uparrow$) & ($\uparrow$) &($\downarrow$) & ($\uparrow$) & ($\uparrow$) & ($\uparrow$) & ($\uparrow$) & ($\downarrow$) & ($\uparrow$)\\
    \midrule
    \textbf{Trained on SPLASH} & 31.15 & 38.26 & 71.03 & 12.30 & 64.72 & 25.70 & 23.23 & 59.86 & 23.23 & 75.14\\
    \textbf{\hspace{5mm}+$\mathcal{D}_{editsql}^{temp}$} & 31.15 & 37.68 & 71.49 & 14.82 & 64.63 & 25.70 & 15.68 & 56.69 & 26.05 & 75.14\\
    \textbf{\hspace{5mm}+$\mathcal{D}_{editsql}^{sim}$ (ours)} & \textbf{33.10} & \textbf{41.60} & \textbf{74.14} & \textbf{11.49} & \textbf{65.45} & \textbf{29.22} & \textbf{23.99} & \textbf{61.97} & \textbf{19.71} & \textbf{76.11}\\
    \bottomrule
    \end{tabular}
    }
    \caption{Error correction performance (\%) on the SPLASH and EditSQL test sets, when the model is trained on the original SPLASH training set, and optionally augmented by the template feedback ($\mathcal{D}_{editsql}^{temp}$) or our simulated feedback ($\mathcal{D}_{editsql}^{sim}$) based on EditSQL's mistakes on the Spider training set. ($\uparrow$: higher, better; $\downarrow$: lower, better)
    }
    \label{tab:results}
\end{table*}

\subsection{Experimental Setup}\label{sec:exp-setup}
We conduct experiments using the SPLASH dataset \cite{elgohary-etal-2020-speak}, which contains human-annotated feedback for mistakes made by the Seq2Struct parser \cite{shinSeq2Struct} on the Spider text-to-SQL semantic parsing dataset \cite{yu-etal-2018-spider}. Specifically, both the SPLASH training (6,829 examples) and dev (810 examples) set were derived from the Spider training set, and the SPLASH test set (870 examples) was from the Spider dev set.\footnote{In our data preprocessing, we removed examples which require adding or removing an entire subquery, since human feedback to these errors is very noisy. We provide details in Appendix~\ref{app:data-preprocess}.}

\vspace{1mm}
\noindent \textbf{Experimental Settings.} To demonstrate the effectiveness of our feedback simulator and evaluator, we consider two settings:

\noindent \textbf{(1) Simulating feedback to a specific semantic parser:}
We investigate whether our feedback simulator trained on the SPLASH dataset can simulate feedback for an unseen semantic parser. In experiments, we follow \citet{elgohary-etal-2020-speak} and experiment with the EditSQL parser \cite{zhang-etal-2019-editing}. 
Specifically, we first follow a similar procedure of \citet{elgohary-etal-2020-speak} to create mistakes made by EditSQL on the Spider training set, and then apply our feedback simulator to simulate NL feedback. This results in around 2,400 simulated training examples. This data is then used to augment the original SPLASH training set for training an error correction model. 
We evaluate the error correction model on both the SPLASH test set and the EditSQL test set (which similarly contains human-annotated feedback to EditSQL's mistakes on the Spider dev set and was additionally provided by \citet{elgohary-etal-2020-speak}).

In this setting, we compare three variants of the error correction model (to be introduced shortly).
\textbf{(a) Trained on SPLASH}, where the model is trained using the original SPLASH training set; \textbf{(b) Trained on SPLASH + $\mathcal{D}_{editsql}^{sim}$}, where the model is trained on both the SPLASH training set and our simulated feedback based on EditSQL; \textbf{(c) Trained on SPLASH + $\mathcal{D}_{editsql}^{temp}$}, where, instead of using our simulated feedback, we use the \emph{template feedback} to augment the training, following the spirit of \citet{yao2019interactive, elgohary-etal-2021-nl}.

\vspace{1mm}
\noindent \textbf{(2) Simulating feedback in low-data settings:}
One important motivation of our research is to reduce the need for human annotations. Therefore, we also experiment with a ``low data'' setting, where only $K\%$ of the SPLASH training set will be used to construct our feedback simulator and evaluator. For the remaining (100$-K$)\% of training examples, we will instead apply our feedback simulator to simulate NL feedback. In experiments, we consider $K$=20, 10, and 5, consuming 1639, 836, and 268 training examples, respectively. Similar to setting (1), we compare our simulated feedback with the template feedback, and will demonstrate the effectiveness of our feedback simulator by evaluating the error correction model trained on its simulation.\footnote{Potentially, one can also apply the simulator to EditSQL for data augmentation, like in setting (1). Here, we focus on solely the low-data setting for easier model comparison.}

For both experiments, we use the TQES feedback simulator variant as it presents the best generation quality, as we will discuss in Section~\ref{subsec:exp_feedback_sim}.
We also note that our proposed feedback evaluator is only used for comparing and selecting better feedback simulator checkpoints or variants. In the future, one can further use our evaluator to provide reward signals when training the feedback simulator (see a discussion in the Limitations section).

\vspace{1mm}
\noindent \textbf{Error Correction Model Evaluation.}
We follow \citet{elgohary-etal-2021-nl} in using four evaluation metrics to assess an error correction model.
\textbf{Correction Accuracy} measures the exact set match \cite{yu-etal-2018-spider}\footnote{The original exact set match does not consider the literal values in a SQL query, but we take it into account because many parsing mistakes involve values.} between the gold parse ($Y^*$) and the parse after correction ($Y_{fix}$).
\textbf{Edit-Dec} and \textbf{Edit-Inc} measure the percentage of test examples for whom the required revision edits are decreased and increased, respectively, after the error correction. Therefore, a better error correction model should expect a larger Edit-Dec but a smaller Edit-Inc. \textbf{Progress} measures the relative edit reduction from revising the corrected vs. initial logical form to the ground truth. Finally, we include the end-to-end (\textbf{E2E}) accuracy of a parser on the Spider dev set, which measures the parsing accuracy when the parser is able to interact with users and correct mistakes via the trained error correction model.

Due to the lack of open-source error correction models, we have implemented our own based on T5 \cite{raffel2020exploring}, with the model details included in Appendix~\ref{app:subsec:error_correction}. While improving the base error correction model is outside our scope, we empirically show that our T5-based error correction model obtains comparable performance to the existing models. We include the comparison and all implementation details in Appendix~\ref{app:sec:impl}.

\subsection{Can the Feedback Simulator Generate Useful Feedback for a Specific Parser?}\label{subsec:exp_full_data}
In Table~\ref{tab:results}, we report results for the experimental setting (1), comparing the performance of different error correction model variants when they are trained using our simulated feedback on EditSQL's mistakes or not. As shown in the table, when including our simulated feedback, we are able to improve the error correction performance for EditSQL by 3.5\% absolute correction accuracy. Note that the correction accuracy is a very strict metric counting only \emph{fully correct} logical forms. On other metrics based on \emph{partial corrections}, we observe that including our simulated feedback can improve them by 5-8\%. These improvements imply that our feedback simulator is able to simulate high-quality NL feedback for errors present in EditSQL (but may be infrequent in SPLASH), which allows the error correction model to better fit EditSQL's test-time error patterns. We present an example in Appendix~\ref{app:subsec:feedback_sim_example}.

We also show that including the simulated feedback on EditSQL can improve the error correction for Seq2Struct (i.e., on the SPLASH test set) as well; it leads to around 2\% gain on correction accuracy and 2.5-3.5\% on others. It is plausible that these gains are not as large as those on the EditSQL test set, given that the additional feedback is simulated based on EditSQL.

Intriguingly, our results present a negative impact from the template feedback. Training the error correction model additionally on the template feedback on EditSQL causes either no gain in Correction Accuracy and worse performance on Progress, especially on the EditSQL test set. Our conjecture is that adding template feedback that describes errors differently from real users can only hinder the error correction model from understanding natural feedback in this full data setting (we will discuss its different impact in low-data settings in Section~\ref{subsec:exp_low_data}).

Finally, looking at the end task accuracy, we note that for both Seq2Struct (the base parser of SPLASH) and EditSQL, being able to correct test-time mistakes based on user NL feedback offers them parsing performance comparable with state-of-the-art parsers on the Spider benchmark. Training their error correction models on our simulated feedback leads to 1\% further gain.

\begin{table}[t!]
    \centering
    \resizebox{0.75\columnwidth}{!}{%
    \begin{tabular}{l|cccc}
    \toprule
    \textbf{Metrics} & \textbf{MRR (dev)} & \textbf{Human} \\
    \midrule
    \textbf{BLEU} & 0.57 & 0.03\\
    \textbf{BERTScore} & 0.55 & 0.08\\
    \textbf{Our Evaluator} & \textbf{0.88} & \textbf{0.19}\\
    \bottomrule
    \end{tabular}
    }
    \caption{Performance of different feedback evaluation metrics. \textbf{MRR} shows the evaluator performance when it is used to rank positive feedback on SPLASH-dev (higher, better). \textbf{Human} denotes their Spearman ranking correlations with human ratings.}
    \vspace{-5mm}
    \label{tab:feedback_eval}
\end{table}

\begin{figure*}[t!]
    \centering
    \includegraphics[width=\linewidth]{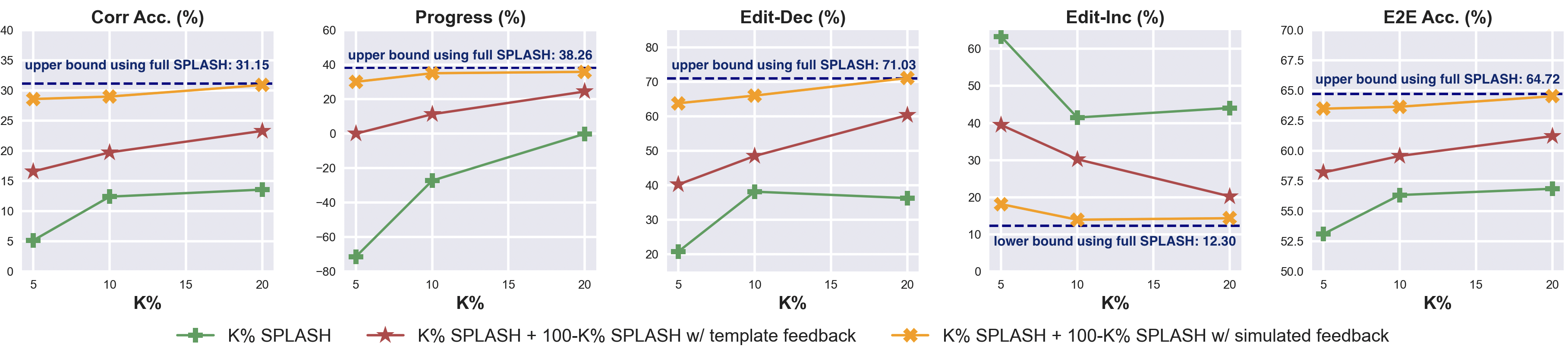}
    \caption{Error correction performance in low-data settings, where only K\% of the SPLASH training set is used and the remaining is simulated using our simulator or the templates. The performance is compared to the upper (lower) bound that was trained using the full SPLASH train set.}
    \label{fig:low-data-results}
\end{figure*}

\subsection{Can the Feedback Evaluator Properly Assess Each Simulator?}\label{subsec:exp_feedback_eval}
As described in Section~\ref{sec:exp-setup}, we rely on our feedback evaluator to select the best feedback simulator. As a result, it is critical that our feedback evaluator can give us precise comparisons across different simulators. We conducted two evaluations comparing our evaluator with the existing metrics, BLEU \cite{papineni-etal-2002-bleu} and BERTScore \cite{zhang2019bertscore}. For automatic evaluation, we report the Mean Reciprocal Rank (MRR) of each evaluation metric when it is used to rank the positive feedback among the 50 negative ones on the SPLASH dev set; the higher MRR, the better metric. 
In addition, we performed a human evaluation and instructed human participants to rank among feedback generated by different simulators under the same context. We then calculate the Spearman ranking correlation between the rank by each evaluation metric and that by humans. We include more human evaluation details in Appendix~\ref{app:subsec:human_evaluation}.

We present the results in Table~\ref{tab:feedback_eval}. On both metrics, our feedback evaluator substantially outperforms the other two metrics. It demonstrates that our evaluator can more precisely assess the logical consistency of a simulated feedback sentence and distinguish between feedback with good and bad quality. In contrast, BERTScore tends to give high values to all generated feedback as long as they are relevant, as we showcase in Appendix~\ref{app:subsec:eval_comp}.

\subsection{How Does Each Feedback Simulator Variant Perform?}\label{subsec:exp_feedback_sim}
We compare the performance of the three feedback simulators (Section~\ref{subsec:feedback_gen}) in Table~\ref{tab:feedback_sim}. While we present performance using different evaluation metrics, as discussed previously, the results of BLEU and BERTScore are relatively less reliable. Results from our evaluator show that TQES can achieve the best performance. We conjecture that this is owing to two advantages. First, compared with CWQES, which requires inferring the desired edits from the incorrect and the correct logical form, TQES directly includes the edit information as input, which simplifies the feedback simulation problem. Second, while both DQES and TQES include the edit information in the input, TQES additionally translates the information into texts, which fits better with how the T5 model was pre-trained (i.e., on textual data). Therefore, in all our experiments, we have been using the TQES-based feedback simulator by default.

\begin{table}[t]
    \centering
    \resizebox{0.85\columnwidth}{!}{%
    \begin{tabular}{l|cccc}
    \toprule
    \textbf{Model} & \textbf{BLEU} & \textbf{BERTScore} & \textbf{Our Evaluator} \\
    \midrule
    \textbf{CWQES} & 0.132 & 0.881 & 0.491 \\
    \textbf{DQES} & \textbf{0.134} & 0.882 & 0.518 \\
    \textbf{TQES} & 0.125 & \textbf{0.884} & \textbf{0.535} \\
    \bottomrule
    \end{tabular}
    }
    \caption{Performance of different feedback simulators.
    }
    \vspace{-5mm}
    \label{tab:feedback_sim}
\end{table}

\subsection{Can the Feedback Simulator Work Well in the Low-data Setting?}\label{subsec:exp_low_data}
Finally, we investigate the performance of our feedback simulator and evaluator in the low-data setting. Our results are shown in Figure~\ref{fig:low-data-results}. A surprising finding is that even when trained with only a small amount of training data, our feedback simulator can still generate high-quality feedback that makes the performance of the error correction model comparable to that of using the full SPLASH training set. As we include more human annotations (i.e., from 5\% to 10\% or 20\%), the feedback simulator can generate better feedback, leading to an upward trend in the error correction performance. Unlike in the full-data experimental setting (Section~\ref{subsec:exp_full_data}), when there is only a limited amount of human annotations, including template feedback assists the error correction model training, although the gains are smaller than that of our simulated feedback.
To further understand the feedback simulator performance, in Appendix~\ref{app:subsec:sim_low_data}, we show the performance of low-data feedback simulators using our feedback evaluator. Our results demonstrate that even when the simulator is trained with a small amount of training data, it can still achieve comparable performance to that trained with full SPLASH data.
\section{Related Work}

\paragraph{Interactive Semantic Parsing.}
Motivated by the need to further enhance its performance in practice, \emph{interactive semantic parsing} emerged as a promising solution \cite{wang-etal-2016-learning-language, chaurasia-mooney-2017-dialog, gur-etal-2018-dialsql, su2018natural, labutov-etal-2018-learning, yao2019interactive, yao-etal-2019-model, staniek2021error, yao-etal-2020-imitation, li-etal-2020-mean, zeng-etal-2020-photon, elgohary-etal-2020-speak, mo-etal-2022-towards}. Among others, \citet{gur-etal-2018-dialsql} and \citet{yao-etal-2019-model} explained components in the generated logical form and, if they were wrong, requested users to select the correct ones as feedback. \citet{li-etal-2020-mean} identified uncertain tokens in the language command and requested user choices on their paraphrases for clarification. While the multi-choice feedback was shown to work well for correcting errors in semantic parsing, it suffers from the obvious drawbacks of being less user-friendly and inefficient, as users can only passively respond to the system-presented choices. 

\citet{labutov-etal-2018-learning} and \citet{elgohary-etal-2020-speak} have driven the research a step forward by introducing \emph{NL feedback}.
Particularly, \citet{elgohary-etal-2020-speak} annotated the SPLASH feedback dataset and showed that an error correction model can learn to fix parsing mistakes from NL feedback. In \cite{elgohary-etal-2021-nl}, the authors further investigated a more advanced error correction model, which predicts the \emph{edits} rather than the \emph{corrected logical form} based on NL feedback.
Our work is complementary to the existing effort. Instead of improving the error correction model architecture, we focus on \emph{simulating NL feedback} to reduce the need for human annotations for training the error correction model. When constructing our feedback simulator, we also explore the use of ``edits'' to improve the model performance.

\paragraph{General NLP Research with Human Feedback.}
There is also work outside semantic parsing exploring human feedback for NLP model development \cite{hancock-etal-2019-learning, kreutzer-riezler-2019-self, sreedhar-etal-2020-learning, madaan2021improving, li-etal-2022-using}. For example, \citet{hancock-etal-2019-learning} explored chatbots that can ask for user feedback when the user shows to be unsatisfied with the conversation. In their work, the feedback can often be viewed as human-labeled responses. \citet{li-etal-2022-using} requested human feedback in the form of ratings and explanations for improving retrieval-based question answering. More recently, \citet{ouyang2022training} collected expert rankings of model outputs for fine-tuning GPT-3. Unlike the prior work, we focus on \emph{(corrective) NL feedback}, a type of feedback that is still largely under-explored. While investigating how to improve a semantic parser from NL feedback is out of our scope, it can be an important future topic.
Finally, concurrent to our work, we noticed an increasing interest in refining large language models with NL feedback from the models themselves \cite{chen2023teaching, madaan2023self, kim2023language}. We envision that models' self-refinement and learning from external human feedback can be two complementary directions and their strengths should be leveraged simultaneously. We will leave the exploration of this topic to the future.

\paragraph{User Simulation in Dialogue Systems.}
User simulation has also been studied with task-oriented dialogue systems \cite{li2016user, shi-etal-2019-build, mohapatra-etal-2021-simulated-chats, kim-etal-2021-neuralwoz}. There, a user simulator typically simulates not only the user utterances but also their goal (e.g., booking a movie ticket at 8pm this Saturday) and their ``agenda'' \cite{schatzmann2009hidden} toward accomplishing the task (e.g., what information to present in the user's first and second conversation turns). Compared with the prior research, our work targets a very different setting, i.e., simulating NL feedback toward correcting the parsing mistakes. We focus this work on developing feedback simulators that can effectively simulate the feedback (i.e., utterance generation), whereas leaving other dimensions of user simulation (e.g., the agenda of error correction) to the future.

\paragraph{Text Evaluation.}
Finally, our work relates to research on text evaluation. Similar to prior work \cite{sulem-etal-2018-bleu, zhang2019bertscore, sellam-etal-2020-bleurt}, in our experiments, we also observe that metrics based on the surface form of a text, such as BLEU \cite{papineni-etal-2002-bleu}, cannot recognize semantic modifications in text generation. Recent research has thus shifted to neural network-based text evaluation, exemplified by metrics such as BERTScore \cite{zhang2019bertscore}, BARTScore \cite{yuan2021bartscore}, CTC Score \cite{deng-etal-2021-compression}, etc. However, while these metrics work well for general-purpose text evaluation (e.g., checking the similarity between two translations), empirically we found them unable to identify the differences between two texts at the more subtle logical level. Therefore, we instead train a text evaluation model for assessing the simulated feedback sentence, following the same spirit of \citet{sellam-etal-2020-bleurt,rei-etal-2020-comet}.
\section{Conclusions}
In this work, we propose the task of simulating NL feedback for interactive semantic parsing and present two models for feedback evaluation and simulation, respectively. Our experimental results have demonstrated the effectiveness of both models and show the promise of saving human-annotation effort with simulated feedback.

\section*{Limitations}\label{sec:limitations}
Both the feedback simulator and the feedback evaluator in our work can be further improved. For example, while we simply fine-tuned a pre-trained T5 model as the feedback simulator, future work can design more specialized architectures for it, such as adding relation-aware attention \cite{wang-etal-2020-rat, elgohary-etal-2021-nl} to augment the schema item linking among input components (e.g., question and template feedback in the TQES variant).
Alternatively, one can also leverage the feedback evaluator to steer the training of the feedback simulator (e.g., via reinforcement learning).
As we briefly discussed, one could also extend our feedback simulator to imitate more fine-grained user behaviors, such as the agenda of how users would engage in the error correction process. Finally, an intriguing research direction is whether one can leverage our feedback simulator for continually improving a semantic parser from NL feedback, drawing inspirations from \citet{clarke-etal-2010-driving, iyer-etal-2017-learning, yao-etal-2020-imitation}.

Although our proposed approaches have not made any assumptions on the type of logical forms and can thus be applied to any of them, in experiments, we have only evaluated them in the task of text-to-SQL semantic parsing. Future research can further assess our proposed models in other semantic parsing settings such as knowledge base question answering \cite{cai2013semantic, yih-etal-2016-value, gu2021beyond, mo-etal-2022-towards}.

On the other hand, as our simulator is primarily designed for interactive semantic parsing, it assumes meaning representations of both the ground-truth prediction and the model prediction. Therefore, generalizing our methods to other NLP tasks may need additional effort. For example, if we apply our methods to a similar interaction scenario for retrieval-based QA \cite{li-etal-2022-using}, then we will additionally need to define logical forms to describe the ground-truth retrieval process and that of the QA model. For open-ended tasks such as keyword-based story generation \cite{pascual-etal-2021-plug-play}, defining such logical forms will need non-trivial effort.

\section*{Ethics Statement}
We presented the task of simulating NL feedback for interactive semantic parsing. The dataset we used in this project is publicly available.  While it is possible that our feedback simulator may generate texts that do not perfectly align with the intended error correction, it is important to note that these generated texts are exclusively used for training the error correction model and are not exposed to real human users. Hence, we do not anticipate any ethical issues resulting from our work.
On the other hand, we emphasize the positive impact of our work when it aims to facilitate feedback-driven human-AI interaction. As shown in this and prior work, human feedback allows for correcting model mistakes before their negative impact takes place, which can play a key role toward enabling safe and trustworthy AI/NLP applications. 

\section*{Acknowledgements}
We would like to thank all anonymous reviewers for their constructive comments. This project was supported by resources provided by the Office of Research Computing at George Mason University (\url{https://orc.gmu.edu}) and funded in part by grants from the National Science Foundation (Awards Number 1625039 and 2018631).

% Entries for the entire Anthology, followed by custom entries
\bibliography{anthology,custom}
\bibliographystyle{acl_natbib}

\appendix
\section{Additional Model Details}

\begin{table*}[h]
    \centering\small
    \resizebox{\linewidth}{!}{%
    \begin{tabular}{|p{70mm}|l|}
    \hline
    \multicolumn{2}{|c|}{\textbf{SELECT Correction}} \\
    \hline
    replace column(s) \textbf{col} (optionally with aggregators \textbf{agg}) & \temp{\textit{primary\{}} find $[\textbf{agg}_{correct}]$ $\textbf{col}_{correct}$ \temp{\textit{\}} \textit{secondary\{}} in place of $[\textbf{agg}_{correct|wrong}]$ $\textbf{col}_{wrong}$ . \temp{\textit{\}}} \\
    \hline 
    add column(s) \textbf{col} (optionally with aggregators \textbf{agg}) & \temp{\textit{primary\{}} additionally find $[\textbf{agg}_{correcy}]$ $\textbf{col}_{correct}$  \temp{\textit{\}}} \\
    \hline
    remove column(s) \textbf{col} (optionally with aggregators \textbf{agg}) & \temp{\textit{primary\{}} do not return $[\textbf{agg}_{wrong}]$ $\textbf{col\_name}_{wrong}$ .\temp{\textit{\}}}\\
    \hline
    add DISTINCT keyword & \temp{\textit{primary\{}} make sure no repetition in the results. \temp{\textit{\}}}\\
    \hline
    remove DISTINCT keyword & \temp{\textit{primary\{}} permit repetitions in the results. \temp{\textit{\}}}\\
    \hline
    \hline
    \multicolumn{2}{|c|}{\textbf{FROM Correction}} \\
    \hline
    replace table(s) \textbf{tab} & \temp{\textit{primary\{}} use $\textbf{tab}_{correct}$ table \temp{\textit{\}}} \temp{\textit{secondary\{}} in place of the $\textbf{tab}_{wrong}$ table.\temp{\textit{\}}} \\
    \hline
    add table(s) \textbf{tab} & \temp{\textit{primary\{}} additionally use the information from the  $\textbf{tab}_{new}$ table \temp{\textit{\}} \textit{secondary\{}} besides the $\textbf{tab}_{old}$ table.\temp{\textit{\}}} \\
    \hline
    \hline
    \multicolumn{2}{|c|}{\textbf{WHERE Correction}} \\
    \hline
    replace condition(s) \textbf{cond} & \temp{\textit{primary\{}} consider the $\textbf{cond}_{correct}$ condition \temp{\textit{\}} \textit{secondary\{}} in place of the $\textbf{cond}_{wrong}$ condition. \temp{\textit{\}}} \\
    \hline
    change the connector AND|OR between conditions & \temp{\textit{primary\{}} you should consider (both | either) of the conditions rather than (either | both) of them. \temp{\textit{\}}}\\
    \hline
    \hline
    \multicolumn{2}{|c|}{\textbf{ORDER BY ... ASC/DESC LIMIT ... Correction}} \\
    \hline
    add a clause or change both column(s) \textbf(col) and the order direction \textbf{order\_dir} & \temp{\textit{primary\{}} order the results $\textbf{order\_dir}_{correct}$ by $\textbf{col}_{correct}$ \temp{\textit{\}}} \temp{\textit{secondary\{}} in place of ordering $\textbf{order\_dir}_{wrong}$ by $\textbf{col}_{correct}$ .\temp{\textit{\}}}\\
    \hline\hline
    \multicolumn{2}{|c|}{\textbf{GROUP BY column HAVING condition Correction}}\\
    \hline
    replace condition operand \textbf{cond\_opd} & \temp{\textit{primary\{}} find the $\textbf{cond\_opd}_{correct}$ \temp{\textit{\}} \textit{secondary\{}} in place of $\textbf{cond\_opd}$.\temp{\textit{\}}} \\
    \hline
    \end{tabular}
    }
    \caption{Examples of template feedback.}
    \label{tab:template}
\end{table*}

\subsection{Template Feedback}\label{app:subsec:template}
The template feedback is used to describe the edits in a more natural way. We use template feedback in both our feedback simulator and evaluator and it brings several advantages as we stated in section \ref{sec:method}. A SQL query can be divided into different clauses and errors vary in a specific clause. We mainly focus on three kinds of operations that can be used to correct the error parse: replace, add, and remove. In Table \ref{tab:template}, we present examples of our template feedback. For ease of presentation, we use \textbf{col\_name} as the placeholder of a real column name in the database. Similarly for other kinds of schema items (e.g., table names, operators, etc.). Besides, we use subscript $_{correct}$ and $_{wrong}$ to indicate the wrong and correct schema item in the replace operation, use subscript $_{new}$ and $_{old}$ to indicate the newly added schema item in add operation, and use numbers as subscript to indicate multiple schema items in one template.

\subsection{Post-processing of Feedback Evaluation}\label{app:subsec:postprocessing}
We observe that the positive candidate typically has one-to-one alignments with the reference. Inspired by \citet{li-etal-2020-mean}, at test time we additionally perform a Bipartite Matching to encourage one-to-one alignments in the matrix $\mathrm{\mathbf{A}}$, before calculating the similarity score.

Furthermore, we noticed that spans in the reference (i.e., template) feedback contribute differently to describing the error correction intent. For example, when a user would like to replace a certain schema item with an alternative one, they will indicate the correct alternative, but may or may not mention the incorrect one (i.e., a user may say ``show \emph{only} the student name'' instead of ``show the student name \emph{and remove student IDs}''). Therefore, when we calculate the similarity score in practice, we additionally weigh the more important spans with a higher weight and the less important ones with fewer. In the template feedback, we split tokens into primary\_span and secondary\_span, and assign them weights $w_{prm}, w_{sec} \in \mathbb{R}$, such that $w_{prm} + w_{sec} = 1 $. For the ease of presentation, we unify these two weights as $w_{span}$. Use $\mathrm{\mathbf{A}^{b}}$ to indicate the alignment matrix with one-to-one alignments after Bipartite matching. The final similarity score is calculated:
\[
    s_{prec}(T,C)=\frac{1}{M \cdot Z^M}\sum_{m=1}^M\max_n\mathrm{\mathbf{A}_{nm}^{b}}\times w_{span},
\]
\[
    s_{rec}(T,C)=\frac{1}{N  \cdot Z^N}\sum_{n=1}^N\max_m\mathrm{\mathbf{A}_{nm}^{b}}\times w_{span},
\]
\[
    s(T,C)=\frac{1}{2}(s_{prec}+s_{rec}).
\]
Here, $Z^M, Z^N$ denote the normalization term due to the span weighing:
\begin{align*}
    Z^M &= w_{prm} \cdot Cnt_{prm}^M + w_{sec} \cdot Cnt_{sec}^M,\\
    Z^N &= w_{prm} \cdot Cnt_{prm}^N + w_{sec} \cdot Cnt_{sec}^N,
\end{align*}
where $Cnt_{prm}^M$ and $Cnt_{sec}^M$ denote the number of tokens that are primary and secondary spans in the reference feedback, respectively, and $Cnt_{prm}^N$ and $Cnt_{sec}^N$ denote the number of tokens in the candidate feedback whose aligned tokens in the reference side are primary and secondary spans, respectively.

In Table \ref{tab:template}, we present the primary and second spans in the template feedback examples.

\subsection{Error Correction Model} \label{app:subsec:error_correction}
The error correction model targets correcting the initial logical form $Y_{init}$ into the gold one $Y^*$ based on the feedback $F$ as well as other relevant information. Prior work has explored approaches such as re-purposing the multi-turn EditSQL semantic parser \cite{zhang-etal-2019-editing} by feeding the feedback as the second-turn user question \cite{elgohary-etal-2020-speak}, or constructing a transformer-based sequence-to-sequence model \cite{elgohary-etal-2021-nl}. However, none of the models are publicly available. In this work, we create our own error correction model by fine-tuning a pre-trained T5 model \cite{raffel2020exploring}. The model takes as input a sequence of feedback $F$, explanation $E$, the initial question $Q$, as well as the contextual information $S$, and is then trained to generate the ground-truth logical form $Y^*$. Investigating more advanced model architectures for error correction is out of our scope, and we leave it as future work.
\section{Additional Implementation Details}\label{app:sec:impl}

\subsection{Implementation Details}\label{app:subsec:impl}
For feedback evaluation, we sampled 50 negative feedback examples for every positive one during training and evaluation. For tuning the hyper-parameters, we experiment with learning rates in \{1e-5, 1e-6, 1e-7, 1e-8\}, $m$ in \{0.1, 0.3, 0.6\}, and $\lambda$ and $\gamma$ in \{1e-1, 1e-3,1e-5\}. The best configuration is: learning rate 1e-8, batch size 64, $m=0.1$, and $\lambda=\gamma=$1e-3 in the loss function. We trained the evaluator for at most 200 epochs. In post-processing, the primary span weight is set to $0.9$. We select the model parameters that achieve the highest MRR on SPLASH dev set. The same set of hyper-parameters is used for both experimental settings. The feedback simulator is based on T5-large, trained with a learning rate 1e-4. We selected the learning rate of our simulator in the range of \{1e-3, 1e-4, 1e-5\} based on its performance on the SPLASH dev set evaluated via our feedback evaluator. We use a batch size of 5 and a maximum of training steps 10,500. 
Training the evaluator and the simulator requires roughly 48 hours and 10 hours using one NVIDIA A100 80GB GPU, respectively. Our model implementation is based on the Hugging Face transformers library\footnote{https://huggingface.co/docs/transformers/index} and PyTorch version 1.10.2.\footnote{https://pytorch.org/} We have only run experiments using one random seed.

\begin{table}[t!]
    \centering\small
    \resizebox{\columnwidth}{!}{%
    \begin{tabular}{p{30mm}|p{10mm}p{10mm}p{10mm}p{10mm}}
    \toprule
    \textbf{Model} & \textbf{Corr Acc. ($\uparrow$)} & \textbf{Progress ($\uparrow$)} & \textbf{Edit-Dec ($\uparrow$)} & \textbf{Edit-Inc ($\downarrow$)} \\
    \midrule
    \textbf{EditSQL+Feedback \cite{elgohary-etal-2020-speak}} & 25.16 & - & - & - \\
    \textbf{NL-Edit \cite{elgohary-etal-2021-nl}} & 41.17 & 36.99 & 72.41 & 16.93 \\
    \textbf{Ours} & 31.15 & 38.26 & 71.03 & 12.30 \\
    \bottomrule
    \end{tabular}
    }
    \caption{The performance (\%) of our error correction model compared with existing ones.}
    \label{tab:error-correction-results}
\end{table}

\begin{table*}[t!]
    \centering
    \resizebox{\linewidth}{!}{%
    \begin{tabular}{p{18mm}l}
    \toprule
    \textbf{Error Type}: & \textbf{missing entire subquery to UNION clause}\\
    Question: & What are the names of all cities and states? \\
    Correct Parse: & SELECT town\_city FROM addresses UNION SELECT state\_province\_county FROM addresses \\
    Wrong Parse: & SELECT town\_city , state\_province\_county FROM addresses\\
    Explanation: & find the town\_city, state\_province\_county of addresses table\\
    Feedback: & The above sentence is incomplete, so could not paraphrase it. \\
    \midrule
    \textbf{Error Type}: & \textbf{missing entire subquery to EXCEPT clause}\\
    Question: & Show the studios that have not produced films with director "Walter Hill".\\
    Correct Parse: & SELECT studio FROM film EXCEPT SELECT studio FROM film WHERE director = "Walter Hill" \\
    Wrong Parse: & SELECT studio FROM film WHERE director ! = "Walter Hill"\\
    Explanation: & find the studio of film table for which director not equals Walter Hill\\
    Feedback: & don't repeat \\
    \midrule
    \textbf{Error Type}: & \textbf{having entirely redundant subquery from WHERE clause}\\
    Question: & Return the hosts of competitions for which the theme is not Aliens?\\
    Correct Parse: & SELECT hosts FROM farm\_competition WHERE theme != "Aliens"\\
    Wrong Parse: & SELECT theme FROM farm\_competition WHERE competition\_id NOT IN ( SELECT theme FROM farm\_competition )\\
    Explanation: & Step 1: find the theme of farm\_competition table, \\
    & Step 2: find the theme of farm\_competition table whose competition\_id not one of the results of step 1 \\
    Feedback: & Add "theme equals to Aliens" in step 1 , Use hosts in place of theme in step 2.\\
    \midrule
    \textbf{Error Type}: & \textbf{having entirely redundant subquery from INTERSECT clause}\\
    Question: & What is the first name of the students who are in age 20 to 25 and living in PHL city?\\
    Correct Parse: & SELECT fname FROM student WHERE city\_code = "PHL" AND age BETWEEN 20 AND 25\\
    Wrong Parse: & SELECT fname FROM student WHERE city\_code = "PHL" INTERSECT SELECT fname FROM student WHERE age < 20\\
    Explanation: & Step 1: find the fname of student table for which city\_code equals PHL,\\
    & Step 2: find the fname of Student table for which age less than 20, \\
    & Step 3: show the rows that are in both the results of step 1 and the results of step 2 \\
    Feedback: & In step 2 , age must be 20 to 25.\\
    \bottomrule
    \end{tabular}
    }
    \caption{The structural errors in SPLASH. Feedback is noisy and inaccurate if there is a need to add or remove the entire subquery.}
    \label{tab:structural-error}
\end{table*}

\subsection{Dataset and Prepossessing} \label{app:data-preprocess}
Our use of the SPLASH dataset is consistent with their intended use, i.e., for scientific research. The dataset is distributed under the CC BY-SA 4.0 license. The dataset is in English. Its feedback came from anonymized crowd workers at Amazon Mechanical Turk. We refer readers to \citet{elgohary-etal-2020-speak} for more details.

We found that human-annotated feedback is typically noisy and inaccurate if the base parser misses or incorrectly predicts the entire subquery in its prediction. Motivated by it, we defined errors that missed the entire subquery or contained the entire wrong subquery in the initial parse as structural errors and showed several examples in Table \ref{tab:structural-error}. We believe that training our feedback simulator and evaluator with those structural error examples does not bring any benefit. Therefore, we filtered them out of our experiments. We found a total of 652, 61, and 92 structural errors in the SPLASH train, dev, and test set separately.

\subsection{Error Correction Model Implementation}\label{app:subsec:err}
Given that existing error correction models are not open-sourced, we implemented our own model based on T5-base, as detailed in Appendix~\ref{app:subsec:error_correction}. We compare our error correction model with existing ones (when all are trained on SPLASH) in Table~\ref{tab:error-correction-results}. Note that EditSQL+Feedback \cite{elgohary-etal-2020-speak} is a model repurposed from EditSQL \cite{zhang-etal-2019-editing}, but it is different and independent from the EditSQL in our main experiments. NL-Edit \cite{elgohary-etal-2021-nl} is the current state-of-the-art model on SPLASH. Both EditSQL+Feedback and NL-Edit are not publicly available, and reproducing them requires non-trivial effort. Therefore, we only include results reported by the authors. 

We observe a 10\% gap between our model and NL-Edit, although their performances are very comparable in all other metrics. This can be due to that Correct Accuracy is a very strict metric; it requires full correction to be counted as ``correct''. However, in practice, we observe that a large portion of human-annotated feedback sentences on SPLASH are noisy (e.g., containing inaccurate information or being incomplete). In such cases, our model can only correct parts of the model mistakes, which leads to worse Correction Accuracy but comparable Progress and Edit percentages (which count partial corrections).
\section{Additional Experimental Results}\label{app:sec:exp}

\begin{table*}[h]
    \centering
    \resizebox{\linewidth}{!}{%
    \begin{tabular}{p{2cm}p{20cm}}
    \toprule
    \multicolumn{2}{l}{\textbf{Error Pattern:} missing DISTINCT in SELECT, missing table in FROM, two errors in WHERE} \\
    \midrule
    \multicolumn{2}{l}{\textbf{Error case in EditSQL-test}}\\
    Question: & What are the different models created by either the car maker General Motors or weighed more than 3500?\\
    Correct Parse: & SELECT DISTINCT t2.model FROM car\_names AS t1 JOIN model\_list AS t2 ON t1.model = t2.model JOIN car\_makers AS t3 \\ 
    & ON t2.maker = t3.id JOIN cars\_data AS t4 ON t1.makeid = t4.id WHERE t3.fullname = "General Motors" OR t4.weight > 3500\\
    Wrong Parse: & SELECT t3.model FROM car\_makers AS t1 JOIN model\_list AS t2 ON t1.id = t2.maker JOIN car\_names AS t3 ON \\
    & t2.model = t3.model WHERE t1.maker = "General Motors" or t1.maker = 3500 \\
    Explanation: & Step 1: for each row in car makers table , find the corresponding rows in model list table and in car names table, \\
    & Step 2: find the car names 's model of the results of step 1 whose car makers 's maker equals General Motors or car makers 's maker equals 3500 \\
    Human Feedback: & Step 1 , Swap car names with cars data Step 2 , Swap second car makers 's maker with cars data 's weight , Ensure Uniqueness.\\
    
    \midrule
    \multicolumn{2}{l}{\textbf{Error case in EditSQL-train with the same error pattern}} \\
    Question: & find the number of actors from Iran who played in "Jim Jarmusch" movies\\
    Correct Parse: & SELECT COUNT ( DISTINCT t1.name  ) FROM cast AS t4 JOIN actor AS t1 ON t4.aid  =  t1.aid JOIN movie AS t5 ON t5.mid = t4.msid JOIN directed\_by AS t2 ON t5.mid = t2.msid \\
    & JOIN director AS t3 ON t3.did = t2.did WHERE t1.nationality = "Iran" AND t3.name = "Jim Jarmusch"\\
    Wrong Parse: & SELECT COUNT (*) FROM actor WHERE nationality = "val1" AND nationality = "val1"\\
    Explanation: & find the number of rows in actor table whose nationality equals dummy value and nationality equals dummy value \\
    Simulated Feedback: & Make sure that actor is from Iran and also use director's name and corresponding movie's name instead of nationality and val1 respectively.\\
    \bottomrule
    \end{tabular}
    }
    \caption{
    An example of an uncommon error pattern in SPLASH. The same error exists in the EditSQL train and test sets. By including EditSQL in the training set of the error correction model, the model is able to fix the parse with this error pattern. EditSQL itself does not predict literal values. We plug values into the wrong parse of EditSQL by randomly picking one from the database content if possible, however, if the initial parse contains the wrong table/column information, we will use dummy values in place of it such as "val1" in above example.
    }
    \label{tab:unseen-error-editsql}
\end{table*}

\subsection{Example of Feedback Simulation}\label{app:subsec:feedback_sim_example}
To better compare the errors in EditSQL and SPLASH, we first define what is error pattern in SPLASH and EditSQL. Error pattern is used to describe the errors for each clause in the initial wrong parse. If there is a need to add new schema item to a clause without removing other schema items, we say this is a missing schema item, otherwise, it is an erroneous schema item. A common error pattern refers to a pattern that appears many times (>10) in SPLASH, and an uncommon error pattern refers to a pattern that appears less than 10 times in SPLASH. In Table~\ref{tab:unseen-error-editsql}, we show feedback simulated by our model when the error is uncommon in SPLASH but present in the EditSQL (simulated) training and test set. By using both SPLASH and EditSQL train sets, the correction model is able to fix uncommon errors in the EditSQL test that cannot be fixed by using SPLASH alone. Even though the simulated feedback is not perfect, we can still see that our feedback simulator generates high-quality feedback for this uncommon error pattern. 
In Table~\ref{tab:evaluation-example}, we also show simulation examples on the SPLASH dataset.

\begin{table*}[t!]
    \centering
    \resizebox{0.99\linewidth}{!}{%
    \begin{tabular}{p{32mm}l}
    \toprule
    \multicolumn{2}{c}{\textbf{Easy Example from SPLASH-dev}} \\
    \midrule
    Question: & How many dogs went through any treatments?\\
    Correct Parse: & SELECT count(DISTINCT dog\_id) FROM treatments\\
    Wrong Parse: & SELECT count ( * ) FROM breeds  \\
    Explanation: & find the number of rows in breeds table\\
    Template Feedback: & use treatments table in place of breeds table . find number of different dog id in place of number of rows . \\
    Human Feedback: & Change breeds table with treatments table . \\
    \midrule
    \multicolumn{2}{c}{\textbf{Simulated Feedback \& Evaluation Results}} \\
    \midrule
    
    CWQES & you need to use treatments table in place of breeds table . \\
    & BLEU: 0.308, BERTScore: 0.876, Ours: 0.468\\
    DQES & you need to use treatments table and search for the number of distinct dog id .\\
    & BLEU: 0.063, BERTScore: 0.879, Ours: 0.528\\
    TQES & you need to use treatments table and also find the number of distinct dog id . \\
    & BLEU: 0.065, BERTScore: 0.889, Ours: 0.529\\

    \bottomrule
    \toprule
    \multicolumn{2}{c}{\textbf{Complex Example from SPLASH-dev}} \\
    \midrule
    Question: & What is the maximum mpg of the cars that had 8 cylinders or that were produced before 1980?\\
    Correct Parse: & SELECT mpg FROM cars\_data WHERE cylinders = 8 OR year < 1980 ORDER BY mpg DESC LIMIT 1\\
    Wrong Parse: & SELECT Max ( T3.horsepower ) FROM model\_list AS T1 JOIN car\_names AS T2 ON T1.model = T2.model \\
    & JOIN cars\_data AS T3 ON T2.makeid = T3.id WHERE T1.model = "amc" OR T3.year < 1  \\
    Explanation: & Step 1: for each row in model list table , find the corresponding rows in car names table and in cars data table \\
    & Step 2: find the maximum horsepower of the results of step 1 whose model list 's model equals amc or year less than 1 \\
    Template Feedback: & in step 1 , do not use the car names table and model list table . in step 2 , consider the year less than 1980 or \\
    & cylinders equals 8 conditions in place of the year less than 1 or model list 's model equals amc conditions . \\
    & find mpg in place of maximum horsepower . find the result with the largest mpg . \\
    Human Feedback: & Change horsepower with mpg , model list's model with cylinders equals 8 and ensure year less than 1980. \\
    \midrule
    \multicolumn{2}{c}{\textbf{Simulated Feedback \& Evaluation Results}} \\
    \midrule
    
    CWQES & replace model list 's model with cars data 's cylinders , replace 1 with 8 , find mpg instead of horsepower , remove or year less than 1 . \\
    & BLEU: 0.062, BERTScore: 0.860, Ours: 0.379\\
    DQES & you should use mpg and cylinders in place of horsepower and model list 's model respectively . also , year should be less than 1980 .\\
    & BLEU: 0.049, BERTScore: 0.869, Ours: 0.410\\
    TQES & find mpg instead of horsepower and ensure that cylinders equals 8 or year less than 1980 . \\
    & BLEU: 0.016, BERTScore: 0.846, Ours: 0.495\\

    \bottomrule
    \end{tabular}
    }
    \caption{
    Two examples show how our evaluator performs compared to BLEU and BERTScore. In both examples, our evaluator correctly ranks all three simulated feedback.
    }
    \label{tab:evaluation-example}
\end{table*}

\subsection{Human Evaluation}\label{app:subsec:human_evaluation}

We conducted a human evaluation to compare different feedback evaluation metrics. Specifically, we randomly sampled 50 examples from the SPLASH dev set, presenting the generated feedback from the three feedback simulators (Section~\ref{subsec:feedback_gen}) but hiding the simulator information, and then asking human participants to score their quality in terms of their logical consistency with the error correction intent. Along with the generated feedback, we also show to participants the question, the correct parse, the wrong parse, the explanation of the wrong parse, and the database schema. We recruited in-house volunteers who are graduate students in computer science.

The human evaluator is working on a 5-level Likert Scale and we include the evaluation criterion showing to human evaluator in Table \ref{tab:human-eval-cri}. For each of the evaluation metrics (i.e., BLEU, BERTScore, and our evaluator), we then calculate the Spearman ranking correlation between the metric values and the human ratings. The reason for using a ranking correlation is that we target an evaluation metric that can help us to distinguish between good and bad feedback simulation under the same context. Intuitively, if a metric can precisely assess different feedback sentences, it should be able to rank these sentences in an order that is similar to the humans'.

\begin{table}[t!]
    \centering\small
    \resizebox{0.9\columnwidth}{!}{%
    \begin{tabular}{p{5mm}p{15mm}|p{50mm}}\toprule
        \multicolumn{2}{l|}{\textbf{Rank}} & \textbf{Description} \\
        \midrule
        1 & \centering{Strongly Disagree} &  The simulated feedback is totally incorrect. (e.g. contains only wrong operations or irrelevant to the edits) \\\midrule
        2 & \centering{Disagree} &  The simulated feedback is partially incorrect. (e.g. contains both wrong and correct operations) \\\midrule
        3 & \centering{Neutral} &  The simulated feedback contains all correct operations, but it is incomplete (partially correct) or contains a lot of (greater and equals 2) unnecessary operations or duplicate operations. \\\midrule
        4 & \centering{Agree} &  The simulated feedback contains correct and complete operations, but it also contains fewer (1) unnecessary operations or duplicate operations. \\\midrule
        5 & \centering{Strongly Agre}e & All operations contained in the simulated feedback are correct, complete, and can be easily followed and understood. There are no additional duplicate operations. \\\bottomrule
    \end{tabular}
    }
    \caption{The human evaluation criterion in a 5-level Likert Scale.}
    \label{tab:human-eval-cri}
\end{table}

\begin{table}[h]
    \centering\small
    \resizebox{0.7\columnwidth}{!}{%
    \begin{tabular}{l|c}
    \toprule
    \textbf{Model (TQES)} & \textbf{Our Evaluator} \\
    \midrule
    \textbf{Trained on SPLASH} & 0.535\\
    \textbf{Trained on 20\% SPLASH} & 0.516\\
    \textbf{Trained on 10\% SPLASH} & 0.491 \\
    \textbf{Trained on 5\% SPLASH} &  0.497\\
    \bottomrule
    \end{tabular}
    }
    \caption{Performance of the low-data feedback simulators trained using different amounts of SPLASH. The evaluation is based on our evaluator.}
    \label{tab:low-data-simulator-eval}
\end{table}

\subsection{Case Study of Evaluation Metrics}\label{app:subsec:eval_comp}
\begin{figure*}[t!]
    \centering
    \includegraphics[width=\linewidth]{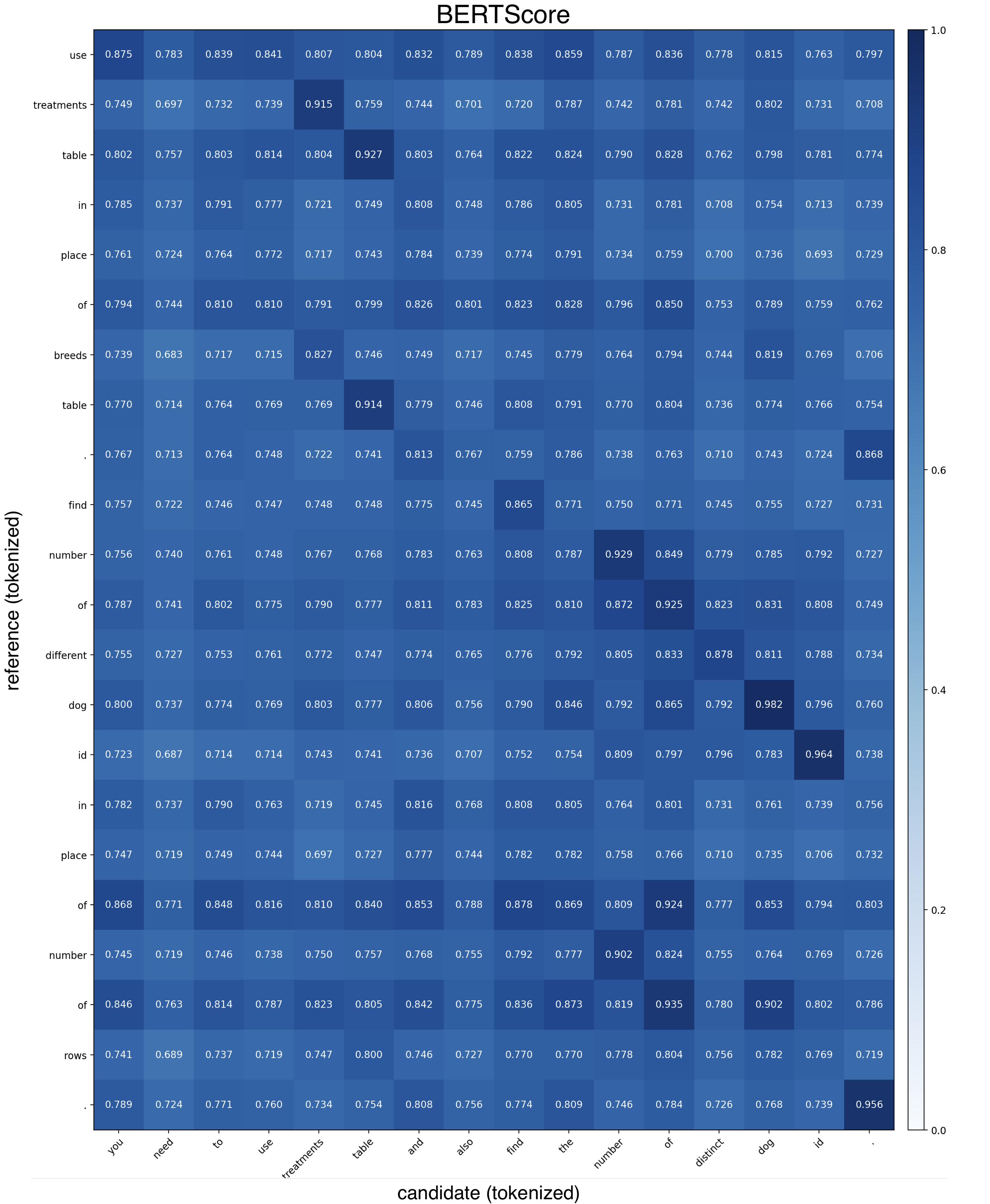}
    \caption{The similarity matrix for easy example shown in Table \ref{tab:evaluation-example} from BERTScore. The candidate simulated feedback comes from the TQES setting.}
    \label{fig:bert-matrix}
    % \vspace{-10pt}
\end{figure*}

\begin{figure*}[t!]
    \centering
    \includegraphics[width=\linewidth]{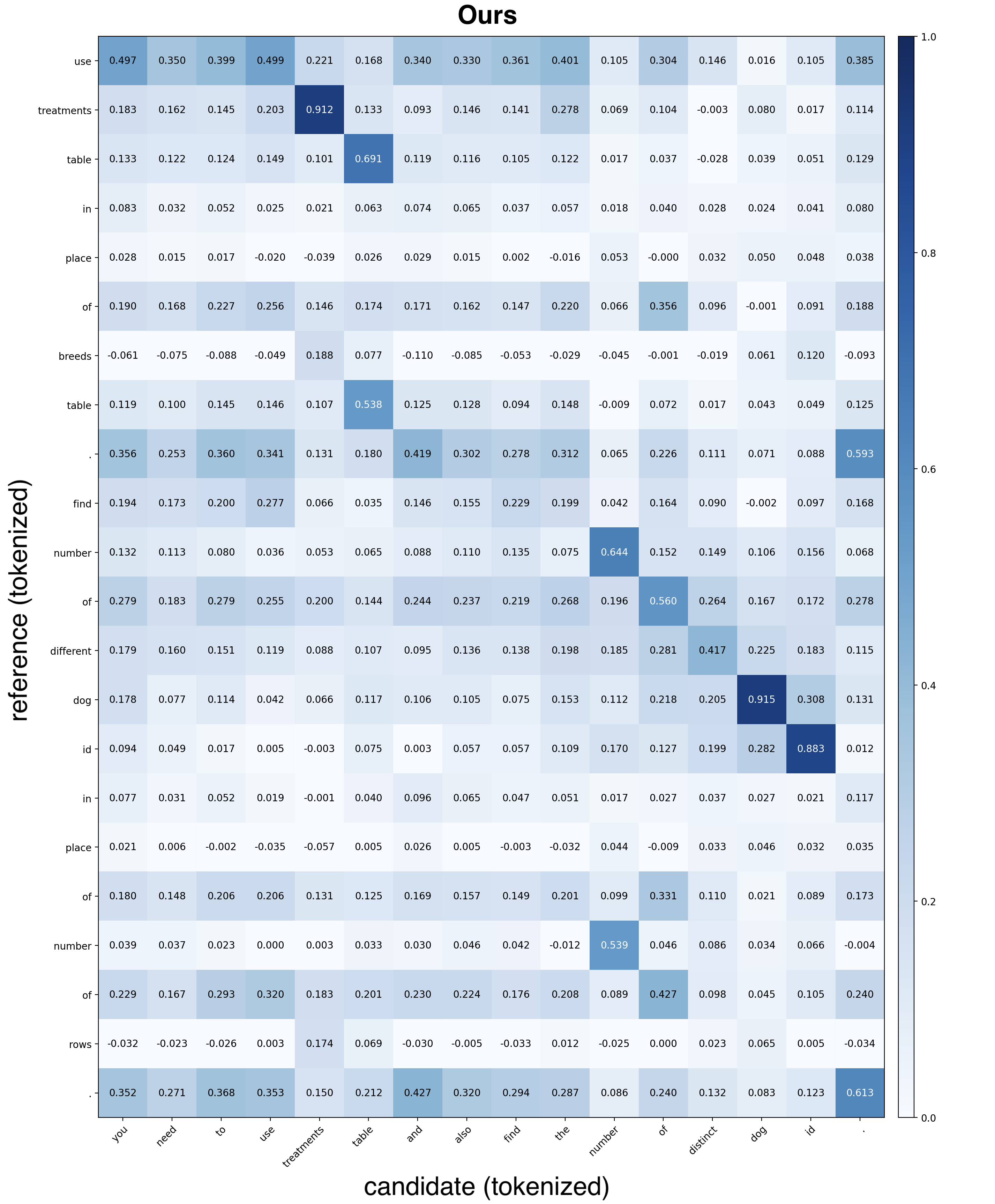}
    \caption{The similarity matrix for easy example shown in Table \ref{tab:evaluation-example} from our evaluator. The candidate simulated feedback comes from the TQES setting.}
    \label{fig:eval-matrix}
    % \vspace{-10pt}
\end{figure*}

In this section, we showcase how our evaluator outperforms BLEU and BERTScore. In Table \ref{tab:evaluation-example}, we included two examples from our feedback simulator and evaluator. In the easy example, our evaluator suggests equally good for DQES and TQES simulated feedback, but BERTScore gives a greater margin between this two simulated feedback and BLEU score incorrectly gives the CWQES the highest score. For the complex example, our evaluator successfully detects the logical inconsistency in CWQES and TQES settings and gives a relatively lower score than TQES, but both BLEU and BERTScore failed to estimate the simulated feedback correctly. Moreover, for both examples, our feedback simulator generates high-quality feedback in the TQES setting. In Figure \ref{fig:bert-matrix} and \ref{fig:eval-matrix}, we show the token-level similarity matrix generated by BERTScore and our evaluator. Our evaluator generates a sparser and more accurate matrix than BERTScore.

\subsection{Feedback Simulation in Low-data Settings}\label{app:subsec:sim_low_data}

In Table~\ref{tab:low-data-simulator-eval}, we evaluate feedback simulators trained in different low-data settings. We evaluate them using our evaluator trained on the full SPLASH; however, we note that in low-data experiments, the feedback evaluator used to select the best simulator was trained consistently using the same small amount of SPLASH data. It is observed that even when we used only 20\% of the SPLASH training data, the learned feedback simulator can still present comparable generation quality, which explains the small gap between error correction models trained using the full SPLASH and with our simulated feedback (Figure~\ref{fig:low-data-results}).

\end{document}